\documentclass[10pt,twocolumn,letterpaper]{article}

\usepackage{cvpr}              
\usepackage{amsmath}
\usepackage{amsthm}
\usepackage{graphicx} 
\usepackage{subcaption}
\usepackage{tabularx}
\usepackage{multirow}
\usepackage{float}
\usepackage{amsthm}

\usepackage[dvipsnames]{xcolor}

\definecolor{cvprblue}{rgb}{0.21,0.49,0.74}
\usepackage[pagebackref,breaklinks,colorlinks,citecolor=cvprblue]{hyperref}

\def\paperID{16692} 
\def\confName{CVPR}
\def\confYear{2024}

\title{Enhancing Multimodal Large Language Models with Multi-instance Visual Prompt Generator for Visual Representation Enrichment}

\author{Wenliang Zhong\\
The University of Texas at Arlington\\
{\tt\small wxz9204@mavs.uta.edu}
\and
Wenyi Wu\\
Amazon\\
{\tt\small wenyiwu@amazon.com}
\and
Qi Li\\
Amazon\\
{\tt\small qlimz@amazon.com}
\and
Rob Barton\\
Amazon\\
{\tt\small rab@amazon.com}
\and
Boxin Du\\
Amazon\\
{\tt\small boxin@amazon.com}
\and
Shioulin Sam\\
Amazon\\
{\tt\small shioulin@amazon.com}
\and
Karim Bouyarmane\\
Amazon\\
{\tt\small bouykari@amazon.com}
\and
Ismail Tutar\\
Amazon\\
{\tt\small ismailt@amazon.com}
\and
Junzhou Huang\\
The University of Texas at Arlington\\
{\tt\small jzhuang@uta.edu}
}

\newtheorem{proposition}{Proposition}

\begin{document}
\maketitle
\begin{abstract}

Multimodal Large Language Models (MLLMs) have achieved SOTA performance in various visual language tasks by fusing the visual representations with LLMs leveraging some visual adapters.
In this paper, we first establish that adapters using query-based Transformers such as Q-former is a simplified Multi-instance Learning method without considering instance heterogeneity/correlation. We then propose a general component termed Multi-instance Visual Prompt Generator (MIVPG) to incorporate enriched visual representations into LLMs by taking advantage of instance correlation between images or patches for the same sample. Quantatitive evaluation on three public vision-language (VL) datasets from different scenarios shows that the proposed MIVPG improves Q-former in main VL tasks. 
\end{abstract}    
\section{Introduction}
\label{sec:intro}

In recent years, with the disruptive changes brought to the Machine Learning community by Large Language Models (LLMs)\cite{radford2018improving, brown2020language, radford2019language, OpenAI2023GPT4TR}, an increasing number of researchers have been exploring the application of LLMs in the realm of multimodality, giving rise to Multimodal Large Language Models (MLLMs)\cite{alayrac2022flamingo, li2023blip, li2022blip, zhu2023minigpt, liu2023visual}. One of the most common forms of multimodality involves the combination of images and text. Just as humans excel in using both images and text to perform tasks, the fusion of images and text in multimodal applications finds wide real-world use, such as in Image Captioning\cite{hossain2019comprehensive, you2016image, yao2017boosting, rennie2017self} and Visual Question Answering (VQA)\cite{antol2015vqa, goyal2017making, wang2017fvqa, lu2016hierarchical}. Leveraging the formidable generalization capabilities of large models, MLLMs have achieved state-of-the-art (SOTA) performance in various few-shot and fine-tuning tasks.
\begin{figure}
    \centering
    \includegraphics[width=\linewidth]{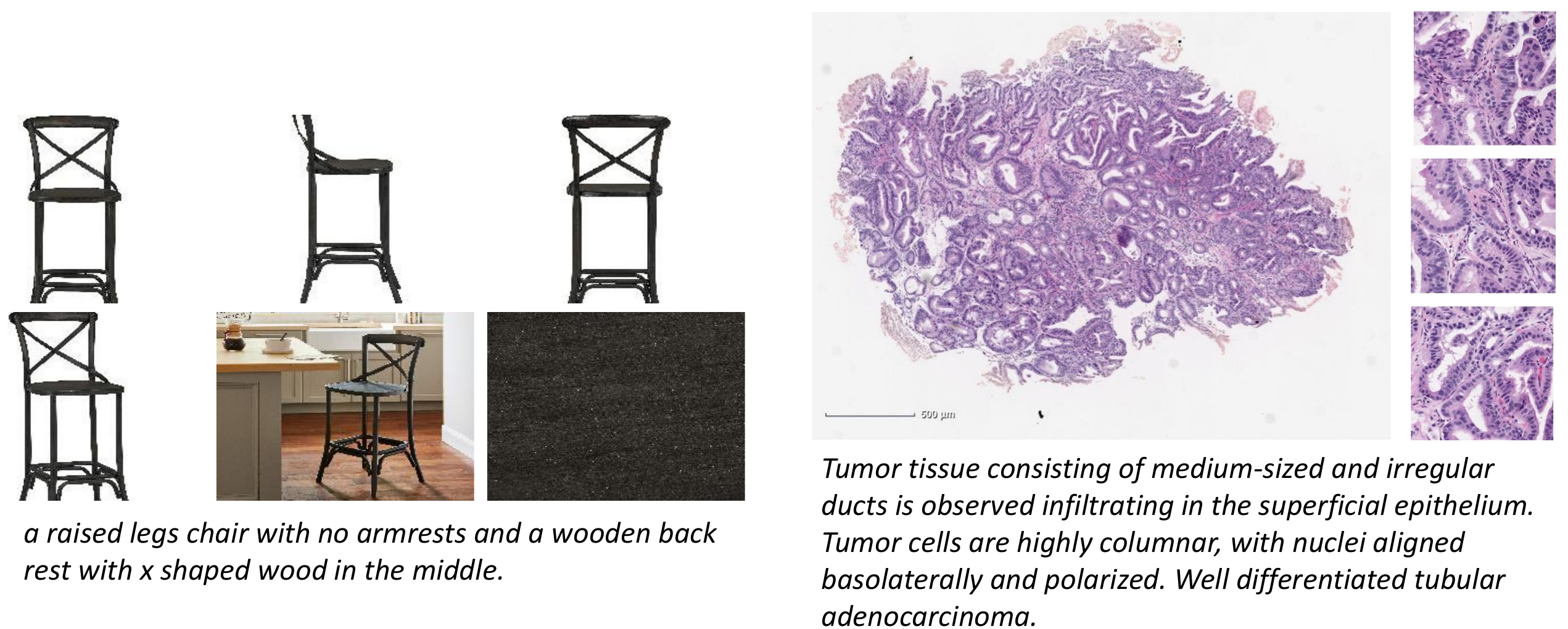}
    \caption{\textbf{Left}: Exemplary images from \cite{collins2022abo}, portraying e-commerce products captured from various aspects. \textbf{Right}: Illustration of a Whole Slide Image (WSI) sourced from \cite{tsuneki2022inference}. Each WSI is composed of multiple patches, exhibiting dimensions comparable to those of natural images.}
    \label{fig:data_overview}
    \vspace{-0.4cm}
\end{figure}

Although MLLMs have achieved remarkable results in various multimodal tasks, the majority of existing open-source MLLMs are primarily pretrained on (image, text) pairs. However, in real-life scenarios, samples are usually represented by enriched visual representations. For example, e-commerce stores typically display products accompanied by several images and a textual description of their features \cite{collins2022abo}. These images may represent different angles of the product's appearance or various aspects of its overall and detailed characteristics. In medical image analysis, a Whole Slide Image (WSI)\cite{tsuneki2022inference} is challenging to fit entirely into a network due to its gigapixel size (more than $10^8$ pixels). Existing medical image analysis typically segments them into multiple patches as images\cite{zhu2017wsisa, ilse2018attention, aeffner2019introduction, li2021dual, zhang2022dtfd}, yet these multiple patches still represent the same sample. Therefore, applying MLLMs to multimodal tasks with richer visual inputs holds much practical significance.

In contemporary MLLMs, the integration of images is achieved through a critical component for imparting visual understanding to LLMs through transforming images to visual tokens, which we termed \textbf{Visual Prompt Generators} (VPGs) in this paper. SOTA MLLMs, such as BLIP2\cite{li2023blip}, Flamingo\cite{alayrac2022flamingo}, and MiniGPT-4\cite{zhu2023minigpt}, utilize attention-based VPGs with learnable query embeddings. These embeddings engage in cross-attention with visual embeddings, extracting visual information for LLM input. In this work, we introduce a novel approach, the Multi-instance Visual Prompt Generator (MIVPG), designed to handle diverse visual inputs. Drawing inspiration from Multiple Instance Learning (MIL), MIVPG treats images or patches of a sample as a set of instances, forming a "bag." Unlike traditional machine learning tasks, MIL performs predictions at the bag level rather than the instance level, employing permutation-invariant functions to aggregate instances. MIVPG extends this concept by considering correlations and relationships across visual representations, facilitating signal pooling from different dimensions. Additionally, we establish that the commonly used QFormer\cite{li2023blip, zhu2023minigpt} is a limited MIL module, prompting the introduction of MIVPG. We showcase MIVPG's enhanced performance across three distinct scenarios, including common natural images, gigapixel-sized pathological images, and e-commerce products with multiple images. 

In summary, our contributions in this paper can be outlined as follows:


\begin{itemize}
    \item We introduce a general and flexible component MIVPG to incorporate enriched visual representations and their relationship into the open source LLM. 
    \item We establish that the commonly used QFormer is a simplified case of MIVPG with limited capability and conduct experiments to show the superiority of our component over the QFormer.
    \item We evaluate the MIVPG on three public datasets from distinct scenarios and showcase that the MIVPG  supports visual representation aggregation from different dimensions: image dimension for e-commerce data and patch dimension for WSI. MIVPG outperforms the QFormer by a significant margin in all datasets, which demonstrates the effectiveness and generalizability of the proposed component. 
\end{itemize}

\section{Related Work}
\subsection{Multimodal Learning}


Recently, various vision-language models (VLMs) have been proposed to enhance the fusion of text and images. For example, TCL
\cite{yang2022vision}
employed triplet contrastive learning to simultaneously learn from text and images. 
Many state-of-the-art MLLMs have also emerged, with one major distinction lying in the design of VPGs. For instance, FROMAGe \cite{koh2023grounding} and LLaVA \cite{liu2023visual} employ a straightforward linear projection as their VPGs. On the other hand, Flamingo \cite{alayrac2022flamingo} introduces the novel use of the Perceiver Resampler, incorporating cross attention and learnable query embeddings. BLIP2 \cite{li2023blip} innovatively employs the QFormer to improve image-text alignment. Meanwhile, MiniGPT-4 \cite{zhu2023minigpt} integrates a frozen QFormer with additional learnable layers for enhanced performance.

While successful in diverse tasks, current multimodal models are primarily designed under the assumption of a one-to-one relationship between texts and image inputs. In reality, the relationship between text and images can be one-to-many or many-to-many. Effectively applying multimodal models in such scenarios poses an open challenge.

\subsection{Multiple Instance Learning}
Traditionally, Multiple Instance Learning
\cite{carbonneau2018multiple,maron1997framework}
can be broadly categorized into two main types: (1) \textbf{The instance-level approach}
\cite{campanella2019clinical, feng2017deep, hou2016patch, kanavati2020weakly}
: In this approach, bag-level predictions are directly derived from the set of instance-level predictions. (2) \textbf{The embedding-level approach}
\cite{ilse2018attention,li2021dual,lu2021data,shao2021transmil}
: Here, bag-level predictions are generated from an bag-level embedding that represents multiple instances. For the former, hand-crafted pooling operators such as mean pooling or max pooling are often employed. However, in practical applications, these hand-crafted pooling operators often yield limited results. Hence, the majority of current research is grounded in the latter approach.

Aggregating instance features to form bag-level features typically leads to better outcomes but requires more complex pooling operations. Recent research has applied neural networks to the pooling process in MIL. For instance, MI-Net
\cite{wang2018revisiting}
utilizes a fully connected layer in MIL. Furthermore, AB-MIL
\cite{ilse2018attention}
employs attention during the pooling process, allowing for better weighting of different instances. Another category of methods\cite{shao2021transmil} attempts to consider the relationships between different instances using the self-attention mechanism. Moreover, DS-MIL
\cite{li2021dual}
employs attention not only to consider instance-to-instance relationships but also instance-to-bag relationships; DTFD-MIL
\cite{zhang2022dtfd}
incorporates the Grad-CAM\cite{selvaraju2017grad} mechanism into MIL. 
While these approaches concentrate on single modality, the extension of MIL to multimodal applications is scarcely explored \cite{wang2023using}.

\section{Methodology}
\label{sec:methodology}

\subsection{Preliminaries and Notations}
\begin{figure*}
    \centering
    \begin{subfigure}[t]{0.7\textwidth}
        \includegraphics[width=\textwidth]{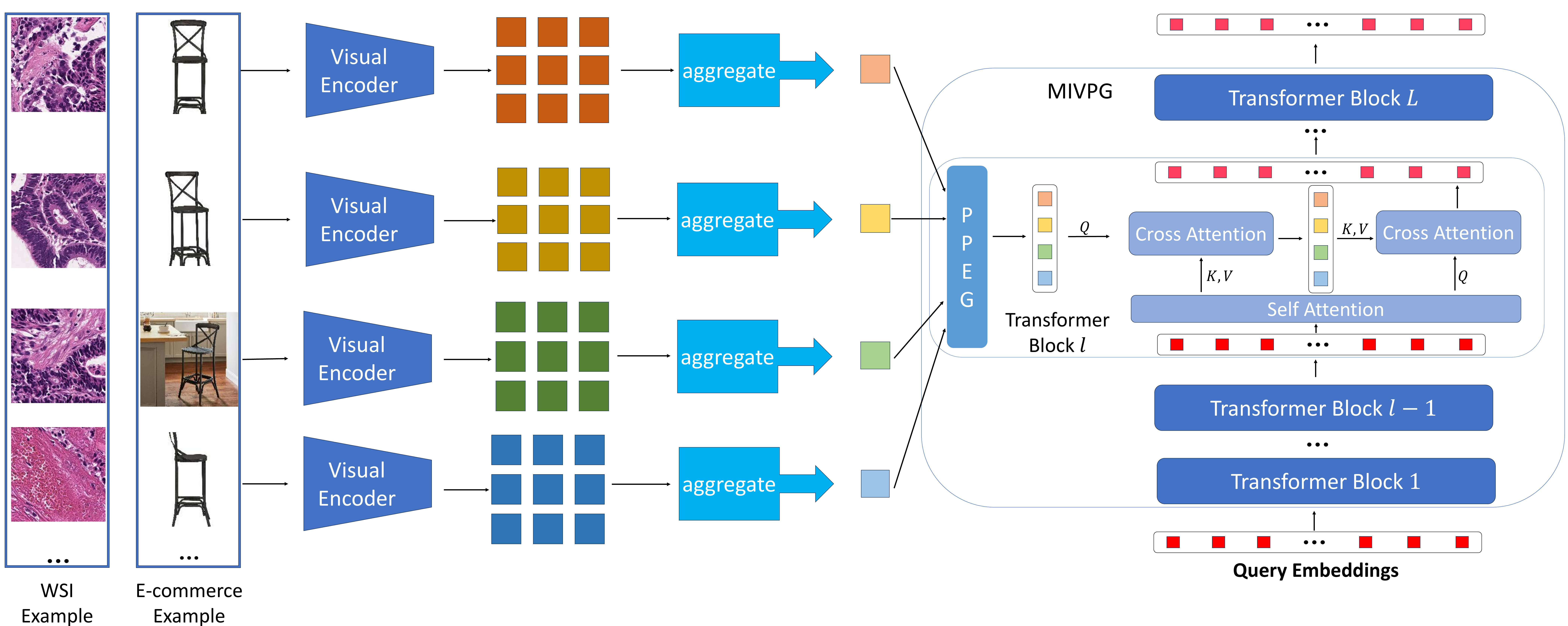}
        \caption{Overview of MIVPG}
        \label{fig:method_overview_1}
    \end{subfigure}
    \\
    \begin{subfigure}[b]{0.4\textwidth}
        \includegraphics[width=\textwidth]{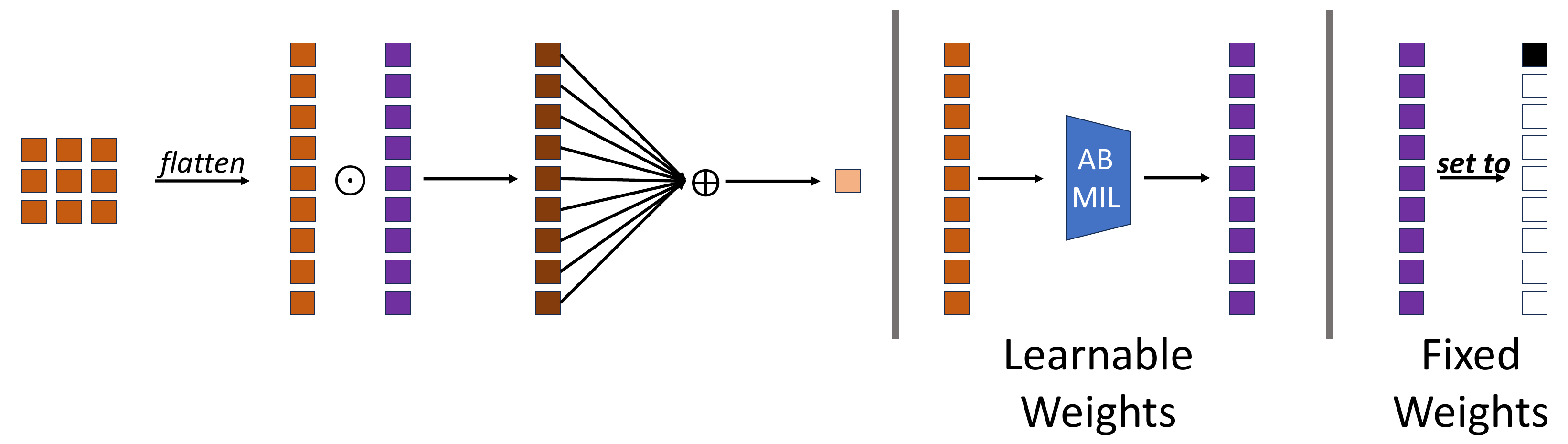}
        \caption{Image-level Bag Aggregation}
        \label{fig:method_overview_2}
    \end{subfigure}
    \hspace{2em}
    \begin{subfigure}[b]{0.4\textwidth}
        \includegraphics[width=\textwidth]{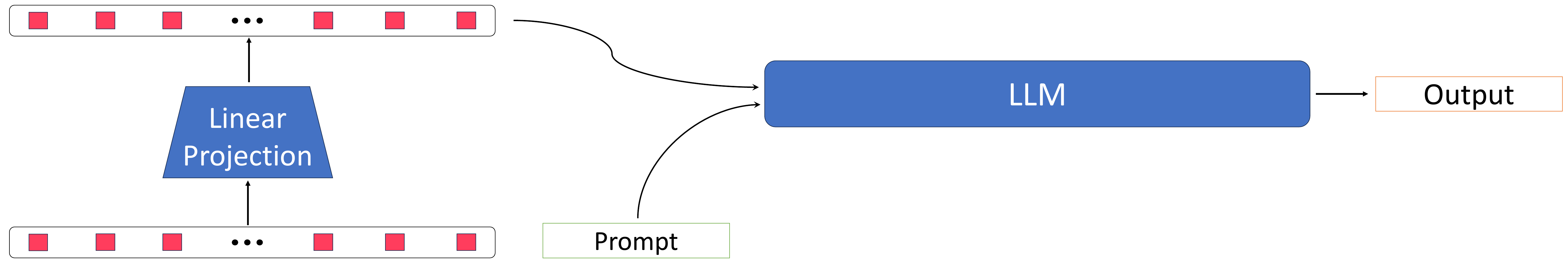}
        \vspace{0.2em}
        \caption{Visual and Text Prompts as Inputs of LLMs}
        \label{fig:method_overview_3}
    \end{subfigure}
    \caption{Overview of MIVPG. \ref{fig:method_overview_1}: When handling multiple visual inputs, the initial step involves aggregating them at the image-level. QFormer can be treated as a Multiple Instance Learning module that takes multiple samples as instances. The MIVPG complements QFormer by introducing a correlated self-attention module and the pyramid positional encoding module, depending on specific scenarios. \ref{fig:method_overview_2}: Image-level aggregation can employ various MIL strategies, either learnable, such as AB-MIL, or fixed, for example, always selecting a specific token. \ref{fig:method_overview_3}: The visual prompt embeddings produced by Q-Former are combined with textual prompt embeddings and forwarded to the LLM for generating outputs.}
    \label{fig:method_overview}
\end{figure*}

Existing MLLMs utilize VPGs, like Qformer and Perceiver Resampler, to encode images to visual tokens. To elaborate, images are initially processed by a visual model, such as ViT\cite{dosovitskiy2020image}, resulting in image embeddings denoted as $I \in \mathbb{R}^{P\times D_I}$, $P$ represents the overall number of visual tokens, encompassing patches and the special token $[CLS]$, and $D_I$ is the dimension of token embeddings. Subsequently, an adapter is employed to project these image embeddings into the LLM embedding space (Figure \ref{fig:method_overview_3}), which we refer to as visual prompt embeddings. The adapter can be as simple as a linear projection. However, a popular choice for contemporary MLLMs is the cross-attention module, where learnable query embeddings interact with image embeddings. We denote query embeddings to have $R$ tokens and $D_q$ dimension $q\in \mathbb{R}^{R\times D_q}$. This mechanism compels the query embeddings to extract essential information from the image embeddings. Without loss of generality, we represent the attention\cite{vaswani2017attention} format as Equation \ref{equ:attention}, and we will use $D$ to denote the dimension of hidden embeddings after projection throughout the rest of this paper.
\begin{align}
    & Q_{\text{result}} = \text{Attention}(Q, K, V) = \text{softmax}(\frac{QK^T}{\sqrt{D}}) V 
    \label{equ:attention}\\
    & \text{where } Q \in \mathbb{R}^{R_1\times D}, K, V \in \mathbb{R}^{R_2\times D}
\end{align}

$A=\text{softmax}(\frac{QK^T}{\sqrt{D}})\in\mathbb{R}^{R_1\times R_2}$ is termed as attention map indicating which entries are pivotal. The cross-attention between query embeddings and image embeddings can be represented as $\text{Attention}(Q=q, K=I, V=I)$. The resulting query embeddings are then forwarded to FeedForward and Residual layers. Specifically, Perceiver Resampler\cite{alayrac2022flamingo} comprises a single cross-attention layer, while the QFormer is a BERT\cite{devlin2018bert} model in which query embeddings interact with image embeddings within each block. We denote $q^{(l)}$ as the query embeddings in the $l^{th}$ block. Details of the QFormer architecture can be found in Appendix \ref{sec:arch_qformer}. Furthermore, when dealing with samples containing multiple images, we use $N$ to denote the number of images $I \in \mathbb{R}^{N\times P\times D_I}$.

MIL typically treats each sample as a bag containing multiple instances. Therefore, we describe MIL in a general form, where a bag is denoted as $B = \{x_1, x_2, …, x_M\}$, with $M$ representing the number of instances in the bag and $x_i$ representing the latent embedding of an instance. In the subsequent discussions in this paper, we will explore MIL from different dimensions thus $x$ and $B$ may have different semantically meanings depending on circumstances. For instance, at the sample level, $M=N$, where instances represent images. At the image level, $M=P$, with instances representing patches.

\subsection{Relations between Attention-based VPG and MIL}
Embedding-level MIL, as a subset of set models, fundamentally relies on the essential property of permutation invariance\cite{ilse2018attention} (Equation \ref{equ:permutation_invariance}) within the aggregation function $g: \mathbb{R}^{M\times D} \rightarrow \mathbb{R}^{D}$.
\begin{equation}
    g(x_1, x_2, \ldots, x_M) = g(x_{\pi(1)}, x_{\pi(2)}, \ldots, x_{\pi(M)})
    \label{equ:permutation_invariance}
\end{equation}

$\{\pi(1), \cdots, \pi(M)\}$ represents a permuted order of the original set.

One popular choice of aggregation is the weighted pooling as shown in Figure \ref{fig:method_overview_2}. In a bag $B = \{x_1, x_2, …, x_M\}$, instance embeddings are transformed into their corresponding weights $\{\alpha_i\}_{i=1}^{M}$ through a nonlinear projection. All instance embeddings are subsequently pooled using the normalized weights to obtain the bag-level representation (Equation \ref{equ:ab_mil}).
\begin{equation}
    x_B = g(x_1, x_2, \ldots, x_M) = \sum_{i=1}^{M} \alpha_i x_i
    \label{equ:ab_mil}
\end{equation}
In AB-MIL\cite{ilse2018attention}, weights are calculated as Equation \ref{equ:abmil_weights}.
\begin{equation}
    \alpha_i = \frac{e^{w^T \tanh(ux_i^T)}}{\sum_{j=1}^{M}e^{w^T \tanh(ux_j^T)}}, w\in \mathbb{R}^{L\times 1}, u \in \mathbb{R}^{L\times D} 
    \label{equ:abmil_weights}
\end{equation}

DeepSets\cite{zaheer2017deep} has demonstrated that when a model comprises a series of permutation-equivalence layers, it is possible to retain permutation invariance in the output. Subsequently, Set Transformer\cite{lee2019set}'s findings confirm that Transformers equipped with self-attention inherently adhere to these principles. A permutation-equivalence layer can be formulated as Equation \ref{equ:permutation_equivalence}.
\begin{equation}
    f_i(x, \{x_1, \cdots, x_M\}) = \sigma_i(\lambda x + \gamma pool(\{x_1, \cdots, x_M\}))
    \label{equ:permutation_equivalence}
\end{equation}

where $pool$ is an aggregation function, $\lambda$, $\gamma$ are learnable scalars, and $\sigma_i$ is the activation function.

\begin{proposition}
    QFormer belongs to the category of Multiple Instance Learning modules.
    \label{pro:QFormer}
\end{proposition}




Within the cross-attention layer of QFormer, every query token computes weights for image embeddings. Query embeddings, being learnable parameters, can be seen as a linear transformation from an instance to its weight. To provide further clarification, each row in the attention map $A$ signifies the weights assigned to instances for aggregation. Consequently, the cross-attention between the learnable query embeddings and the input is permutation invariance.

The result of cross-attention is combined with the original query embeddings using a residual connection. This process can be expressed as shown in Equation \ref{equ:permutation_equivalence}, by replacing $pool$ with Equation \ref{equ:attention}, and setting $\lambda=\gamma=\mathbb{I}$, as illustrated in Equation \ref{equ:permutation_equivalence_cross_attention}, which is permutation equivalence.
\begin{equation}
    f_i(q, I) = q + \text{Attention}(Q=q, K=I, V=I)
    \label{equ:permutation_equivalence_cross_attention}
\end{equation}
Considering that the self-attention layer within the QFormer block adheres to the principles of permutation equivalence, we can conceptualize the QFormer as a multi-head MIL mechanism.

\begin{figure*}
    \centering
    \includegraphics[width=1.\linewidth]{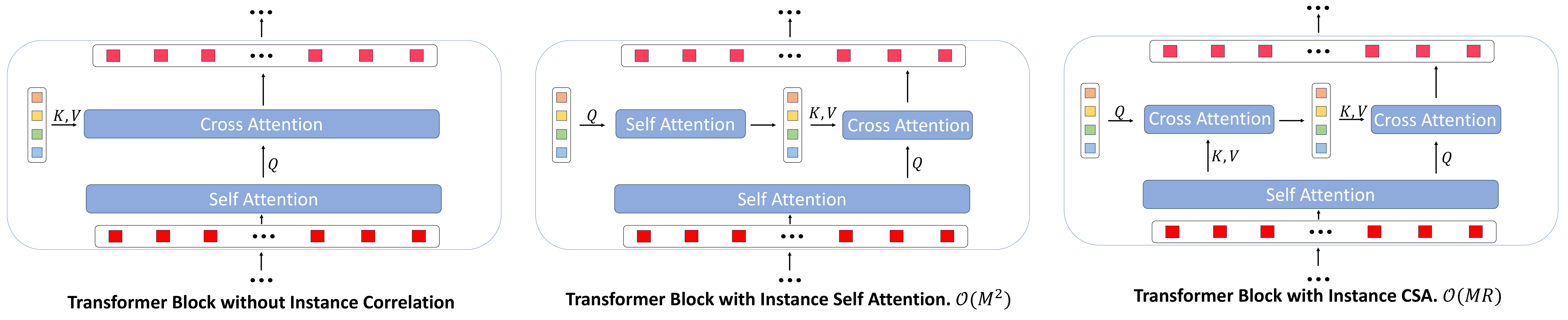}
    \caption{\textbf{Left}: The original transformer block without considering instance correlation. \textbf{Middle}: Instance correlation is computed through a self-attention layer among input instances, incurring a time complexity of $\mathcal{O}(M^2)$. \textbf{Right}: Instance correlation is calculated using query embeddings from the previous layer. This approach reduces the time complexity in computing correlation to $\mathcal{O}(MR)$.}
    \label{fig:method_complexity}
\end{figure*}

From the standpoint of MIL, the weighted pooling in Equation \ref{equ:attention} operates under the assumption that instances are independent and identically distributed (i.i.d)\cite{shao2021transmil}. However, in practical scenarios, instances may exhibit correlations, and accounting for instance correlation can lead to improved performance. It's worth noting that when each sample contains only one image, the input to QFormer comprises patch embeddings that have already incorporated correlations through the self-attention layer in ViT. Moreover, performance enhancement is attainable through the integration of a Pyramid Positional Encoding Generator (PPEG)\cite{shao2021transmil}, which complements the proposed MIVPG when handling single-image inputs.

\subsection{MIVPG for Multiple Visual Inputs}
While previous approaches have touched upon the use of multiple images as inputs to MLLMs, they still exhibit certain limitations. For instance, while handling the scenario of using videos as inputs. Perceiver Resampler\cite{alayrac2022flamingo} simply concatenates patches from multiple images into a sequence to serve as the input, i.e., $I \in \mathbb{R}^{N\times P\times D_I} \rightarrow \mathbb{R}^{(N\cdot P)\times D_I}$. However, it is essential to treat each image as a distinct bag, with each patch considered within the context of its respective bag. Directly flattening the patches may lead to a misallocation of weights across instances from different bags.

When a sample comprises multiple images, it is imperative to consider MIL feature aggregation from different perspectives. In the context of individual images, each image can be treated as a 'bag,' and each patch within the image as an 'instance.' From the sample's perspective, each sample can also be regarded as a 'bag,' with each image within the sample as an 'instance.' When a sample contains only a single image, we can focus primarily on the former perspective since the latter perspective involves a single instance per bag. However, in a more general context, it is essential to adopt a hierarchical approach when considering the utilization of MIL for feature aggregation. Without loss of generality, we now consider the input of the MIVPG to be a bag $B$ containing multiple instances. Hence, the cross-attention can be expressed as $\text{Attention}(Q=q, K=B, V=B)$.

\subsection{Unveiling Instance Correlation in MIVPG for Enhanced Multi-instance Scenarios}

When there are multiple images as inputs and each image is regarded as an instance, their correlation should also be considered. TransMIL\cite{shao2021transmil} has demonstrated that, when multiple instances are as the input, applying self-attention mechanisms can effectively learn the correlations between them. However, in certain applications, such as Whole Slide Image analysis, each bag may contain a large number of instances ($M > 1000$). Directly computing self-attention between instances $\text{Attention}(Q=B, K=B, V=B)$ results in a time complexity of $\mathcal{O}(M^2)$, making such calculations computationally intensive. To address this issue, it\cite{shao2021transmil} employs the Nystrom approximation of attention\cite{xiong2021nystromformer}. In this paper, following QFormer, we propose a method for computing the correlations between instances using a low-rank projection. This approach only requires the incorporation of a Correlated Self Attention (CSA) module in each Transformer block to achieve the desired results.

Considering the input to a Self Attention module is a bag of instance embeddings $B=[x_1, x_2, \dots, x_M]$ with shape $\mathbb{R}^{M\times D}$, instead of directly computing self-attention, one can adopt a more efficient approach\cite{lee2019set} by projecting the original matrices into a lower-rank space using a learnable matrix $L \in \mathbb{R}^{M'\times D}$, where $M \gg M'$, as illustrated in Equation \ref{equ:proj_msa}.
\begin{equation}
    L' = \text{Attention}(Q=L, K=B, V=B)
    \label{equ:proj_msa}
\end{equation}
\begin{equation}
    B_{result} = \text{Attention}(Q=B, K=L', V=L')
    \label{equ:msa_proj}
\end{equation}

Subsequently, the aggregated low-rank matrix can be reintegrated with the original embeddings, as shown in Equation \ref{equ:msa_proj}. This low-rank projection effectively reduces the time complexity to $\mathcal{O}(MM')$.

Recall that QFormer is a stack of Transformer blocks, with each block consisting of a self-attention layer followed by a cross-attention layer. In the $l^{th}$ block, the input consists of query embeddings from the previous block $q^{(l-1)}$ have already aggregated instance information in the prior block. These query embeddings can be directly utilized as $L'$ in the cross-attention layer (as shown in Equation \ref{equ:csa}) since the self-attention layer retains their permutation equivalence. Therefore, we can efficiently harness the query embeddings from the previous block to learn instance correlations.
\begin{equation}
    (B_{result})^{(l)} = \text{Attention}(Q=B, K=q^{(l-1)}, V=q^{(l-1)})
    \label{equ:csa}
\end{equation}

$(B_{result})^{(l)}$ is the updated instance embeddings that have aggregated instance correlation at layer ${l}$.

\begin{proposition}
    MIVPG, when equipped with the CSA (Correlated Self-Attention) module, continues to fulfill the essential properties of MIL
    \label{pro:csa_QFormer}
\end{proposition}

We prove the proposition \ref{pro:csa_QFormer} in the supplementary  \ref{sec:proof_proposition}.

In summary, as depicted in Figure \ref{fig:method_overview_1}, we establish that QFormer falls under the MIL category and is a specialized instance of our proposed MIVPG. The latter extends to visual inputs with multiple dimensions, accounting for instance correlation.


\section{Experiments}
To assess the effectiveness of our proposed approach, we conduct evaluations across various scenarios: 

\begin{enumerate}
    \item where each sample comprises a single image, and patches are naturally considered as instances;
    \item where each sample includes multiple instances, but we use a general embedding for each image;
    \item where each sample contains multiple images, with each image containing multiple patches.
\end{enumerate}

\subsection{General Setup}
We initialize our model using BLIP2 \cite{li2023blip} with FLAN-T5-XL. MIVPG is initialized with weights from QFormer. The model consists of a frozen language model and a frozen visual model. During training, we only update the MIVPG. The visual encoder, ViT-G, is employed to encode images into patches of embeddings, and the images are resized to dimensions of $224\times 224$. In our experiments, we observed that unfreezing the visual encoder does not lead to additional improvements in datasets with small sizes. Further details can be found in the supplementary \ref{sec:freeze_vit}.

\subsection{Scenario 1: Samples with Single Image}
We start by assessing the performance of our method on common single-image datasets to validate the effectiveness of considering Multiple Instance Learning through the addition of Pyramid Positional Encoding Generator for each layer containing MIVPG. Following the fine-tuning baseline in BLIP2, we choose \textbf{MSCOCO}\cite{lin2014microsoft} as the evaluation dataset and employ the Karpathy validation and testing set split. The original training set contains approximately 560K image-text pairs. Given that most existing MIL methods are tailored for small datasets, we evaluate performance across various sizes of training subsets obtained through random sampling. In this dataset, we treat patches as individual instances, and each sample comprises only one image, indicating that $N=1$.


\begin{figure}
    \centering
    \includegraphics[width=0.8\linewidth]{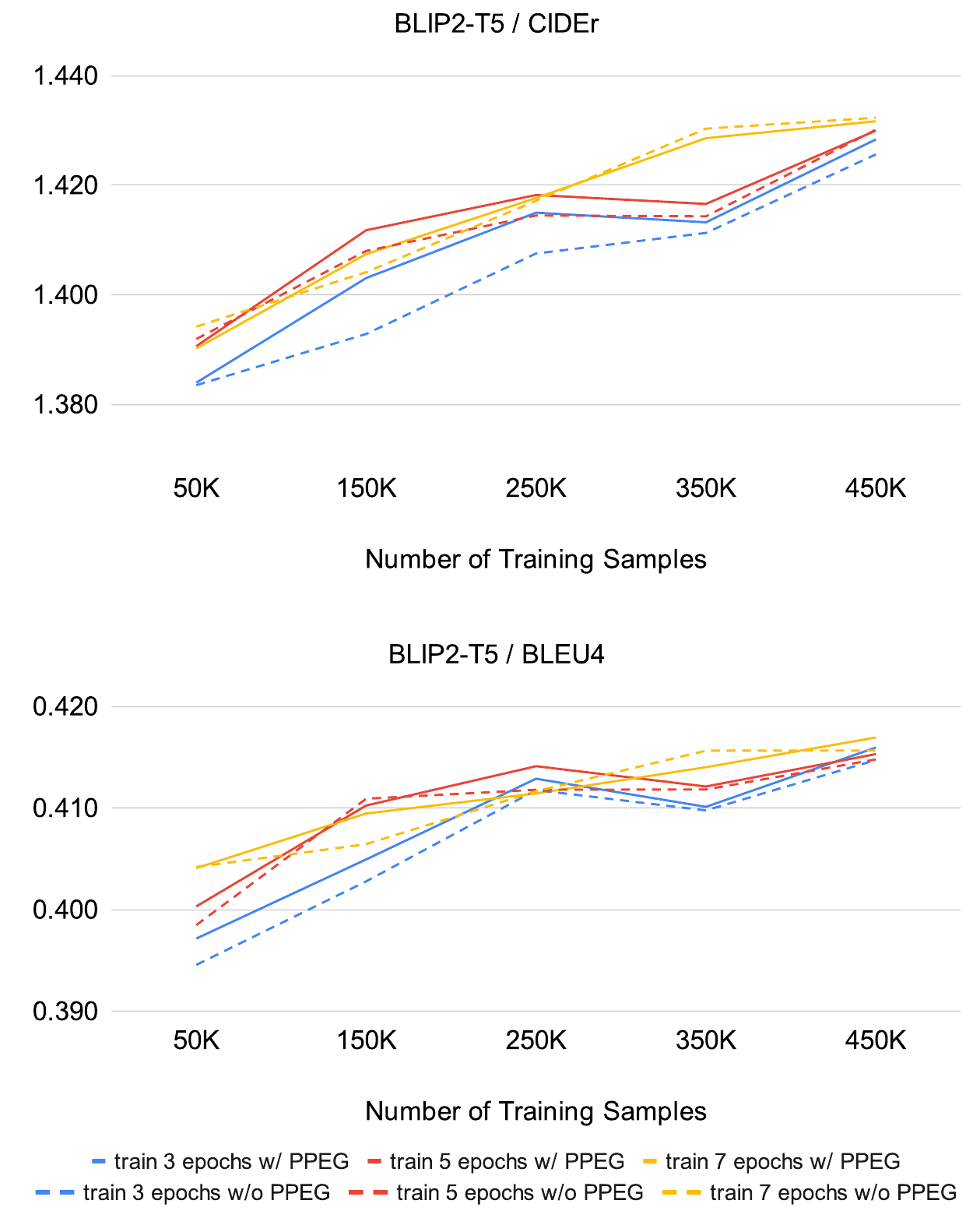}
    \caption{Experiment Results on \textbf{MSCOCO}. We adopt the metrics used in \cite{li2023blip}. It is evident that the incorporation of MIL modules enhances the QFormer in the majority of cases.}
    \label{fig:coco_results}
\end{figure}

The result from the \textbf{MSCOCO} dataset is shown in Figure \ref{fig:coco_results}. It reveals that the enhancements achieved through the use of PPEG are more noticeable when working with smaller datasets. As the dataset size increases, the difference in performance becomes less significant. This can be attributed to the fact that in cases of limited data, models often struggle to discern latent and implicit patterns. Therefore, more sophisticated modules are required to uncover deeper relationships within the data. Conversely, existing MLLMs are typically pretrained on extensive datasets, which tend to mitigate the impact of data scarcity. In practical applications, we demonstrate that one can draw upon MIL techniques to enhance MLLMs performance in scenarios where there is insufficient data for the downstream task.
\begin{table}[]
    \centering
        \caption{Experiments on the PatchGastricADC22 dataset\cite{tsuneki2022inference}, we evaluate our proposed method against baselines from \cite{tsuneki2022inference}, considering four widely-adopted metrics. Augmented baselines, denoted as \textbf{aug}, which signifies a model trained with data augmentation.}
        \label{tab:pga_results}
        \resizebox{0.47\textwidth}{!}{
        \begin{tabular}{ccccc}
            \toprule
             & BLEU@4 & CIDEr & METEOR & ROUGE \\
             \midrule
             DenseNet121 x20 p3x3 & 0.336$\pm 0.023$ & 2.03$\pm 0.245$ & 0.284$\pm 0.009$ & 0.481$\pm 0.016$\\
             EfficientNetB3 x20 p3x3 & 0.364$\pm 0.019$ & 2.154$\pm 0.200$ & 0.302$\pm 0.014$ & 0.510$\pm 0.026$\\
             DenseNet121 x20 p3x3 aug & 0.347$\pm 0.017$ & 2.024$\pm 0.198$ & 0.292$\pm 0.007$ & 0.485$\pm 0.012$\\
             EfficientNetB3 x20 p3x3 aug & 0.414$\pm 0.024$ & 2.820$\pm 0.326$ & 0.327$\pm 0.012$ & 0.540$\pm 0.021$\\
             \midrule
             BLIP2-MIVPG w/o CSA & 0.441$\pm 0.009$ & 2.902$\pm 0.233$ & 0.359$\pm 0.004$ & 0.583$\pm 0.011$\\
             BLIP2-MIVPG (Ours) & \textbf{0.447}$\pm0.012$ & \textbf{2.930}$\pm 0.173$ & \textbf{0.363}$\pm 0.005$ & \textbf{0.590}$\pm 0.004$\\
            \bottomrule
        \end{tabular}}
\end{table}
\vspace{-0.4em}
\subsection{Scenario 2: Samples with Multiple Images, with Each Image as a General Embedding}
Next, we evaluate our method in scenarios involving multiple images, where each image contributes only one embedding as its representation. Specifically, we utilize the \textbf{PatchGastricADC22}\cite{tsuneki2022inference} dataset, which is a Whole Slide Image (WSI) dataset. This dataset includes 991 WSIs of $H\& E$-stained gastric adenocarcinoma specimens, accompanied by diagnostic captions extracted directly from existing medical reports. The dataset encompasses a total of 262,777 medical patches, with each WSI containing up to 1860 patches. Each medical patch has a size of $300\times 300$, which will be encoded by the visual encoder after resizing. The dataset is partitioned into training, validation, and test subsets using the methodology outlined in \cite{tsuneki2022inference}, with a split ratio of 0.7, 0.1, and 0.2, respectively. We compare the proposed method against baselines in \cite{tsuneki2022inference}, which are a combination of a visual model (\textbf{DenseNet121}\cite{huang2017densely} or \textbf{EfficientNetB3}\cite{tan2019efficientnet}) and an LSTM\cite{hochreiter1997long} as the language model. To ensure a fair comparison, we conduct three experiments with different random seeds and follow the same data augmentation in \cite{tsuneki2022inference}. In a medical patch, the focus is typically on global information rather than local details. Additionally, given that a WSI can comprise a large number of patches, we aim to reduce computational overhead. Therefore, we choose to use only the $[CLS]$ token output by ViT as the representation for the entire medical patch. In this case, $P=1$.

\begin{figure*}
    \centering
    \includegraphics[width=0.913\linewidth]{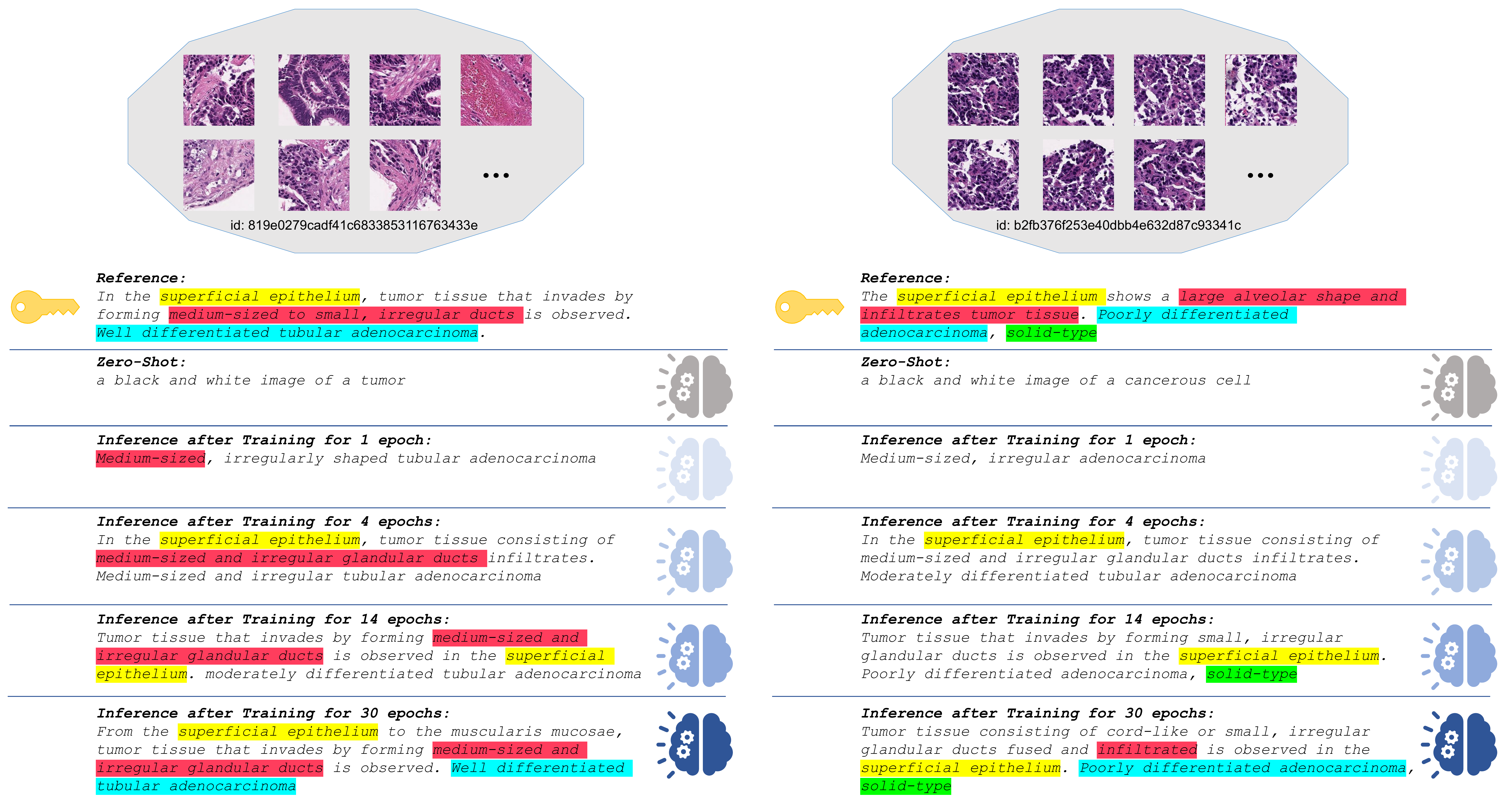}
    \caption{Visualization of Inference Results on PatchGastricADC22. We highlight details that should be focused on the reference. Zero-shot inference is performed using the pretrained BLIP2\cite{li2023blip}. As the number of epochs increases, the model acquires more domain knowledge.}
    \label{fig:pga_visualization}
\end{figure*}
As demonstrated in Table \ref{tab:pga_results}, our method outperforms the baselines significantly. This result highlights the effectiveness of employing large-scale models in downstream tasks. Moreover, the experiments indicate that the model performs even better when considering correlations among instances, underscoring the effectiveness of our CSA module.
\footnotetext[1]{For consistency, we opted for metrics implemented in https://github.com/salaniz/pycocoevalcap.}
Furthermore, we are interested in observing how captions generated by the LLM evolve as the number of training epochs increases. Given the substantial domain gap between medical images and natural images, we believe that existing MLLMs have rarely been trained on medical images, rendering them less domain-specific in medical analysis. As depicted in Figure \ref{fig:pga_visualization}, under the zero-shot setting, BLIP2 struggles to generate detailed captions for the provided WSIs. However, with an increasing number of training epochs, the model acquires domain-specific knowledge and produces more relevant captions. Similar to the process of human learning, a discernible trend is observed, where the model initially generates very general captions and gradually incorporates more and more details as the number of epochs increases.

\subsection{Scenario 3: Samples with Multiple Images, with Each Image Having Multiple Patches to be Considered}
Thirdly, we assess the method in a comprehensive manner where each sample contains multiple images and considers multiple patches within an image. Specifically, we utilize the \textbf{Amazon Berkeley Objects (ABO)}\cite{collins2022abo} dataset, consisting of samples of e-commerce products. Each product is accompanied by multiple images illustrating various characteristics from different perspectives. Typically, one of these images provides an overview of the product, commonly displayed first on an e-commerce website. We refrain from utilizing the product title as the caption due to its limited descriptiveness. Instead, we rely on manually annotated captions \cite{luo2023scalable} from the same dataset as reference captions to assess the quality of our generated captions. Specifically, there are 6410 product samples, and we randomly partition them into train, validation and test subsets with a ratio of 70$\%$, 10$\%$, 20$\%$. Each product comprises images ranging from 2 to 21. In this dataset, we must consider both image details and multiple images simultaneously. Consequently, we apply MIL in both image and patch dimensions. To be specific, we employ AB-MIL (Equation \ref{equ:abmil_weights}) to generate image-level embeddings for images. Each image is then treated as an instance and passed to the QFormer as a sample-level MIL. Since the number of patches per sample is much smaller than that of WSIs, we do not apply PPEG in this setting.
\begin{figure*}
    \centering
    \includegraphics[width=1.\textwidth]{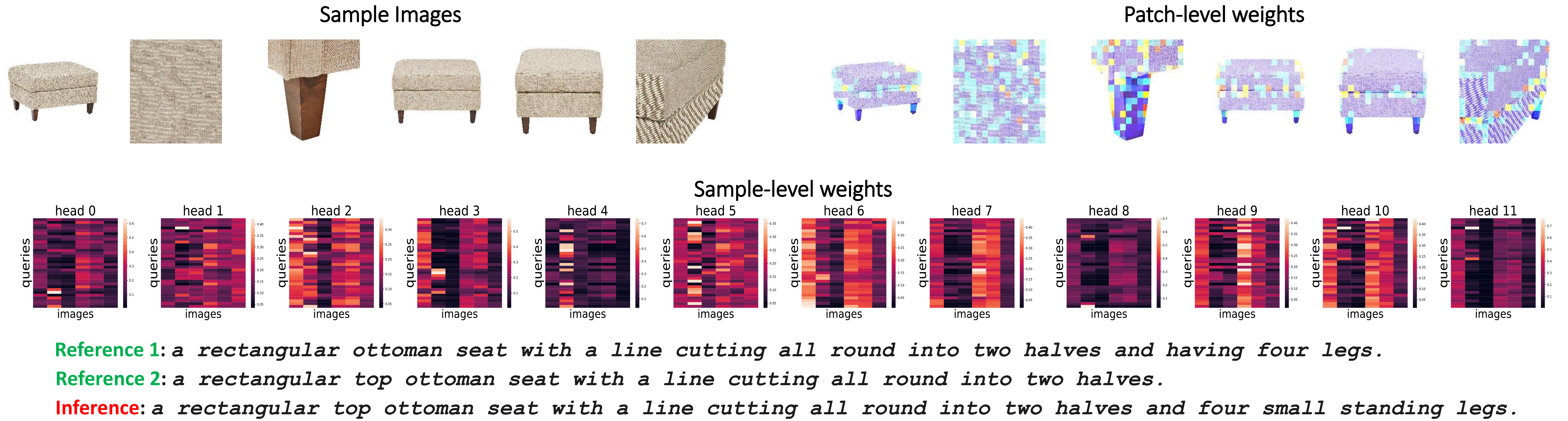}
    \caption{Example from ABO. \textbf{Top Left}: a sample consisting of six images. \textbf{Top Right}: Attention weights of patches among different images. We use the \textit{COLORMAP JET} to represent the weights, where lighter colors indicate higher weights. \textbf{Bottom}: Attention weights among different images. There are 12-head attention maps. In an attention map, each row indicates the weights of images for one query.}
    \label{fig:abo_visualization}
\end{figure*}

We primarily compare the proposed method against BLIP2 with different settings: 1) \textbf{Zero-Shot}: Directly feed the overview image of a sample to query the BLIP2 without fine-tuning; 2) \textbf{Single-Image}: Fine-tune the original BLIP2 with the overview image of each product; 3) \textbf{Patch-Concatenation}: Fine-tune the original BLIP2 with multiple images, with patches concatenated in one sequence.
\begin{table}[!ht]
    \centering
        \caption{Experiment results on the ABO dataset\cite{collins2022abo}. We compare BLIP2 with different settings as our baselines.}
        \label{tab:abo_results}
        \resizebox{0.47\textwidth}{!}{
        \begin{tabular}{ccccc}
            \toprule
             & BLEU@4 & CIDEr & METEOR & ROUGE \\
            \midrule
             BLIP2 Zero-Shot & 0.024$\pm 0.002$ & 0.144$\pm 0.005$ & 0.087$\pm 0.002$ & 0.264$\pm 0.003$\\ 
             BLIP2 Single Image & 0.413$\pm 0.010$ & 1.515$\pm 0.031$ & \textbf{0.304}$\pm 0.006$ & 0.584$\pm 0.006$\\
             BLIP2 Patch-Concatenation & 0.412$\pm 0.011$ & 1.516$\pm 0.032$ & 0.301$\pm 0.006$ & \textbf{0.589}$\pm 0.007$\\
            \midrule
             BLIP2-MIVPG w/o CSA & 0.412$\pm 0.010$ & 1.528$\pm 0.033$ & 0.301$\pm 0.005$ & 0.585$\pm 0.008$\\
             BLIP2-MIVPG & \textbf{0.415}$\pm 0.011$ & \textbf{1.549}$\pm 0.030$ & \textbf{0.304}$\pm 0.005$ & 0.586$\pm 0.008$\\
            \bottomrule
        \end{tabular}}
\end{table}
We repeat the experiments three times with different random seeds and report the mean and standard deviation. As shown in Table \ref{tab:abo_results}, results from the ABO dataset demonstrate that our method outperforms the use of a single image or images with concatenated patches mostly, underscoring the efficacy of considering MIL from different dimensions. 
It is worth noting that fine-tuning BLIP2 with a single image has already achieved respectable performance, indicating that the overview image contains general information about a sample. Additionally, while fine-tuning BLIP2 with multiple concatenated image patches shows good results in terms of \textit{ROUGE}, it should be emphasized that the concatenation results in a complexity of $\mathcal{O}(RNP)$. In contrast, the proposed method applied on different dimensions will only have a complexity of $\mathcal{O}(P + RN)$, ensuring computational efficiency. 

Unlike the groundtruth captions that describe the details of a product, we find the zero-shot BLIP2 tends to provide less detailed information. This discrepancy can be attributed to the model's pretraining, where it is predominantly tasked with describing an overview of an image, with less emphasis on details. Nonetheless, when we input multiple images for a single item, the model showcases its capacity to discern what aspects should be emphasized. This capability arises from the presence of common patterns across different images that collectively describe the item, thus affirming the effectiveness of utilizing multiple visual inputs.

A visualization example can be seen in Figure \ref{fig:abo_visualization}, featuring an ottoman seat composed of six images. We present both patch-level attention weights and image-level attention weights. In the patch-level attention weights, the model emphasizes edges and legs of the seat, leading to an output that recognizes the rectangular shape and four legs. The image-level attention weights show maps for all twelve heads. Each row in a map represents a query, and each column represents an image. Notably, different heads and queries exhibit varying attention patterns towards images. Generally, the first, fourth, and fifth images attract the most attention.


\subsection{Case Study}
\begin{table}[!ht]
    \centering
    \resizebox{0.47\textwidth}{!}{
    \begin{tabular}{|c|c|c|c|c|}
        \hline
         & \multicolumn{2}{c|}{\textbf{ABO}} & \multicolumn{2}{c|}{\textbf{PatchGastricADC22}} \\
        \cline{2-5}
         & \textbf{MIVPG w/ SA} & \textbf{MIVPG w/ CSA} & \textbf{MIVPG w/ SA} & \textbf{MIVPG w/ CSA} \\
        \hline
        \textbf{BLEU@4} & 0.409$\pm 0.012$ & \textbf{0.415}$\pm 0.011$ & 0.444 $\pm 0.020$ & \textbf{0.447} $\pm 0.012$ \\
        \textbf{CIDEr} & 1.532$\pm 0.024$ & \textbf{1.549}$\pm 0.030$ & \textbf{2.961}$\pm 0.242$ & 2.930$\pm 0.173$ \\
        \textbf{METEOR} & 0.299$\pm 0.006$ & \textbf{0.303}$\pm 0.006$ & 0.362$\pm 0.008$ & \textbf{0.363}$\pm 0.005$ \\
        \textbf{ROUGE} & 0.586$\pm 0.008$ & \textbf{0.586}$\pm 0.008$ & 0.586$\pm 0.012$ & \textbf{0.590}$\pm 0.004$ \\
        \hline
    \end{tabular}
    }
    \caption{Ablation results of effectiveness of CSA}
    \label{tab:csa_ab}
\end{table}
To assess the impact of instance correlation, we conduct additional ablation studies involving self-attention (SA) and correlated self-attention (CSA). Please refer to Table \ref{tab:csa_ab} for the results. The results in PatchGastricADC22 indicate that self-attention and correlated self-attention among instances yield similar performance. However, in the case of ABO, correlated self-attention outperforms self-attention. We posit that this discrepancy arises from the fact that images of e-commerce products typically do not exhibit explicit correlations. In the correlated self-attention mechanism, images are initially aggregated into query embeddings, which may help reduce the impact of irrelevant information. Due to space constraints, we have postponed additional case studies to supplementary \ref{sec:more_case_study} and more visualization results to supplementary \ref{sec:more_visualization}.
\section{Conclusion}
In this paper, we introduce the MIVPG, a flexible, general and powerful component to fuse enriched visual representations with MLLMs, which achieves superior performance in diverse use cases. We demonstrate that QFormer is a limited variant of MIVPG, providing theoretical support for its efficacy.We believe the enriched visual signals and advanced MIL-based techniques will contribute to the future development of MLLMs.
{
    \small
    \bibliographystyle{ieeenat_fullname}
    \bibliography{main}
}

\clearpage
\setcounter{page}{1}
\setcounter{section}{0}
\renewcommand{\thesection}{\Alph{section}}
\renewcommand{\thesubsection}{\Alph{section}.\arabic{subsection}}
\maketitlesupplementary

\begin{figure}[!ht]
    \centering
    \includegraphics[width=1.\linewidth]{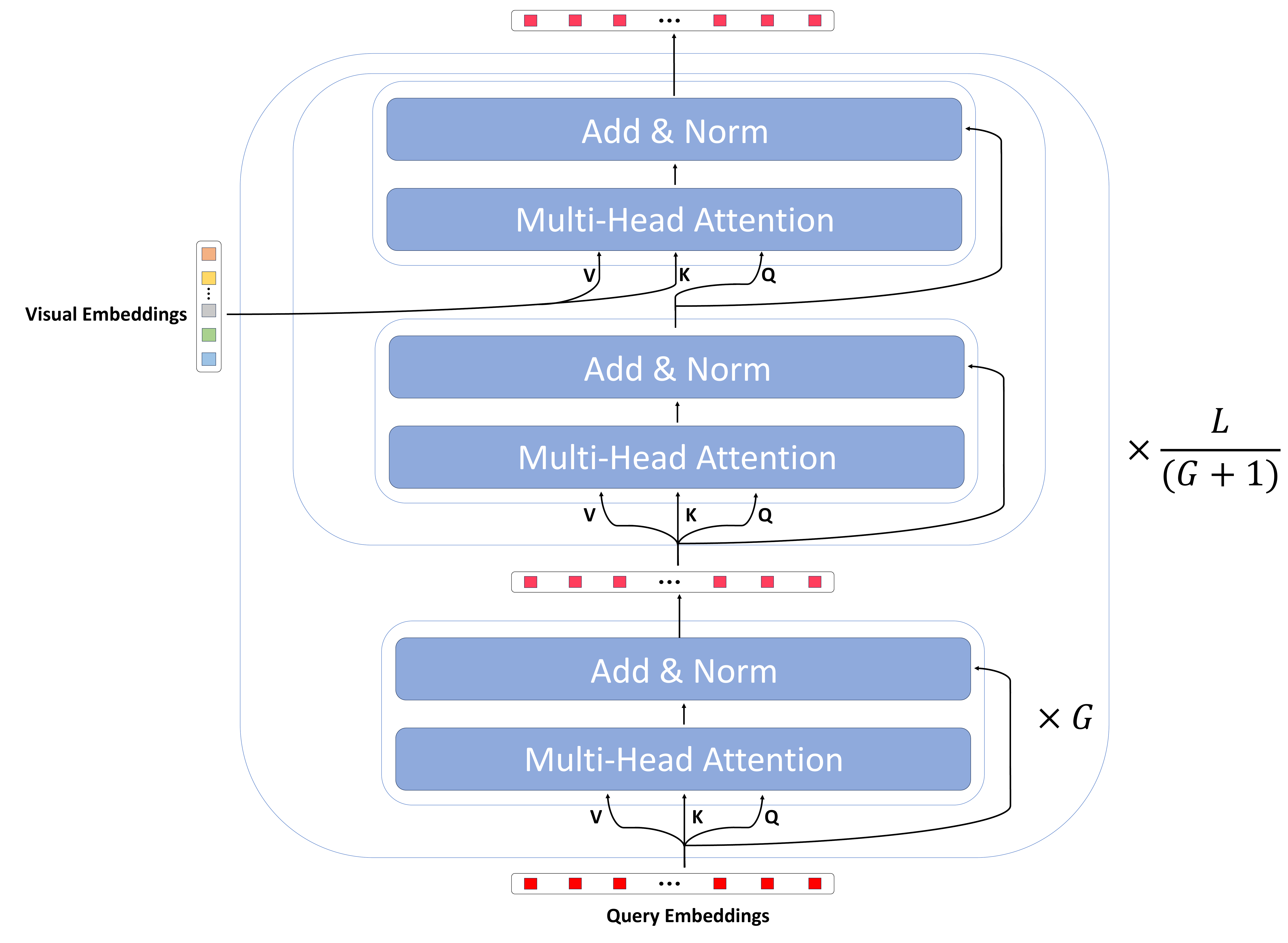}
    \caption{Overview of QFormer}
    \label{fig:QFormer_overview}
\end{figure}

\section{Detailed Architecture of QFormer}
\label{sec:arch_qformer}
The architecture overview is depicted in Figure \ref{fig:QFormer_overview}. Specifically, QFormer is initialized as a BERT-based model\cite{devlin2018bert} comprising a total of $L=12$ layers. In contrast to typical BERT models that process textual inputs, QFormer takes $R=32$ learnable query embeddings as inputs. These embeddings are utilized to extract visual information from the input visual data during Stage-1 pretraining in BLIP2\cite{li2023blip}. Subsequently, they serve as visual prompt embeddings for the LLM inputs after projection.

Inside the QFormer, each layer includes a self-attention module composed of a Multi-Head Attention component and a Forward module (consisting of Linear, LayerNorm, and Residual Connection). The cross-attention module, initialized with random values, is inserted every $G$ layers, where learnable query embeddings interact with visual embeddings.\textit{ In the main paper, for the sake of conciseness, we condensed the representation of the multi-head attention and forward modules into self(cross) attention modules. Furthermore, we exclusively illustrated the modifications made to the cross-attention module in MIVPG, as the self-attention modules remain unchanged.} The final QFormer output is represented by the last layer's query embeddings.

For a more comprehensive understanding, readers are encouraged to refer to \cite{li2023blip}.

\section{Proof of Proposition}
\label{sec:proof_proposition}
In Proposition \ref{pro:csa_QFormer}, we illustrate that MIVPG, when augmented with the CSA (Correlated Self-Attention) module, maintains the crucial permutation invariance property of MIL. In this section, we provide a theoretical demonstration of this property.

\begin{proof}
    Recall that both the original cross-attention and self-attention mechanisms have already demonstrated permutation equivalence for the visual inputs (Property 1 in \cite{lee2019set} and Proposition \ref{pro:QFormer} in the main paper). Our objective is to establish that the CSA module also maintains this permutation equivalence, ensuring that the final query embeddings exhibit permutation invariance. 

A permutation-equivalence function $f$ satisfies the property $\pi (f(x)) = f(\pi (x))$, where $\pi$ is a permutation of the element order in the set $x$. In the context of this paper, we have $f(x=B) = \text{Attention}(Q=B, K=q, V=q)$. Here, we use $B_{\pi}$ to denote the bag after permutation, and $Q_B$, $K_q$, and $V_q$ to denote the $Q, K, V$ matrices in the attention calculation after projection.
\begin{align}
\label{equ:proof_csa}
    f(\pi (x)) &= \text{Attention}(Q=B_{\pi}, K=q, V=q) \\
    &= \text{softmax}(\frac{Q_{B_{\pi}}K_{q}^T}{\sqrt{D}})V_q \\
    &= \pi (\text{softmax}(\frac{Q_{B}K_{q}^T}{\sqrt{D}}))V_q \\
    &= \pi (\text{softmax}(\frac{Q_{B}K_{q}^T}{\sqrt{D}})V_q) \\
    &= \pi (\text{Attention}(Q=B, K=q, V=q)) \\
    &= \pi (f(x))
\end{align}

In Equation \ref{equ:proof_csa}, recall that $\text{softmax}(\frac{Q_{B_{\pi}}K_{q}^T}{\sqrt{D}})$ is the attention map with shape $\mathbb{R}^{M\times R}$ where $M$ is the number of elements in the bag and $R$ is the number of query tokens. The permutation of the bag is row-wise, and therefore $\text{softmax}(\frac{Q_{B_{\pi}}K_{q}^T}{\sqrt{D}}) = \pi (\text{softmax}(\frac{Q_{B}K_{q}^T}{\sqrt{D}}))$. Hence, the CSA module is permutation equivalence. Given that other properties of the original self-attention and cross-attention mentioned above, the MIVPG enhanced by the CSA module remains permutation invariant.
\end{proof}

\section{More Experiments}
We implemented the proposed method on NVIDIA A100 GPUs with BFloat16. Except for the number of training epochs mentioned in the main paper, we kept all other hyperparameters the same as in BLIP2\cite{li2023blip}. For PatchGastricADC22\cite{tsuneki2022inference} and ABO\cite{collins2022abo}, we trained the model for 40 epochs.

\subsection{Frozen Visual Models}
\label{sec:freeze_vit}
In the original BLIP2\cite{li2023blip}, image sizes are upscaled to $364\times 364$, and consequently, the ViT is unfrozen during the fine-tuning process. This approach yields slightly better performance, albeit at a higher computational cost while training on the entire COCO training set.

\begin{figure}
    \centering
    \includegraphics[width=1.\linewidth]{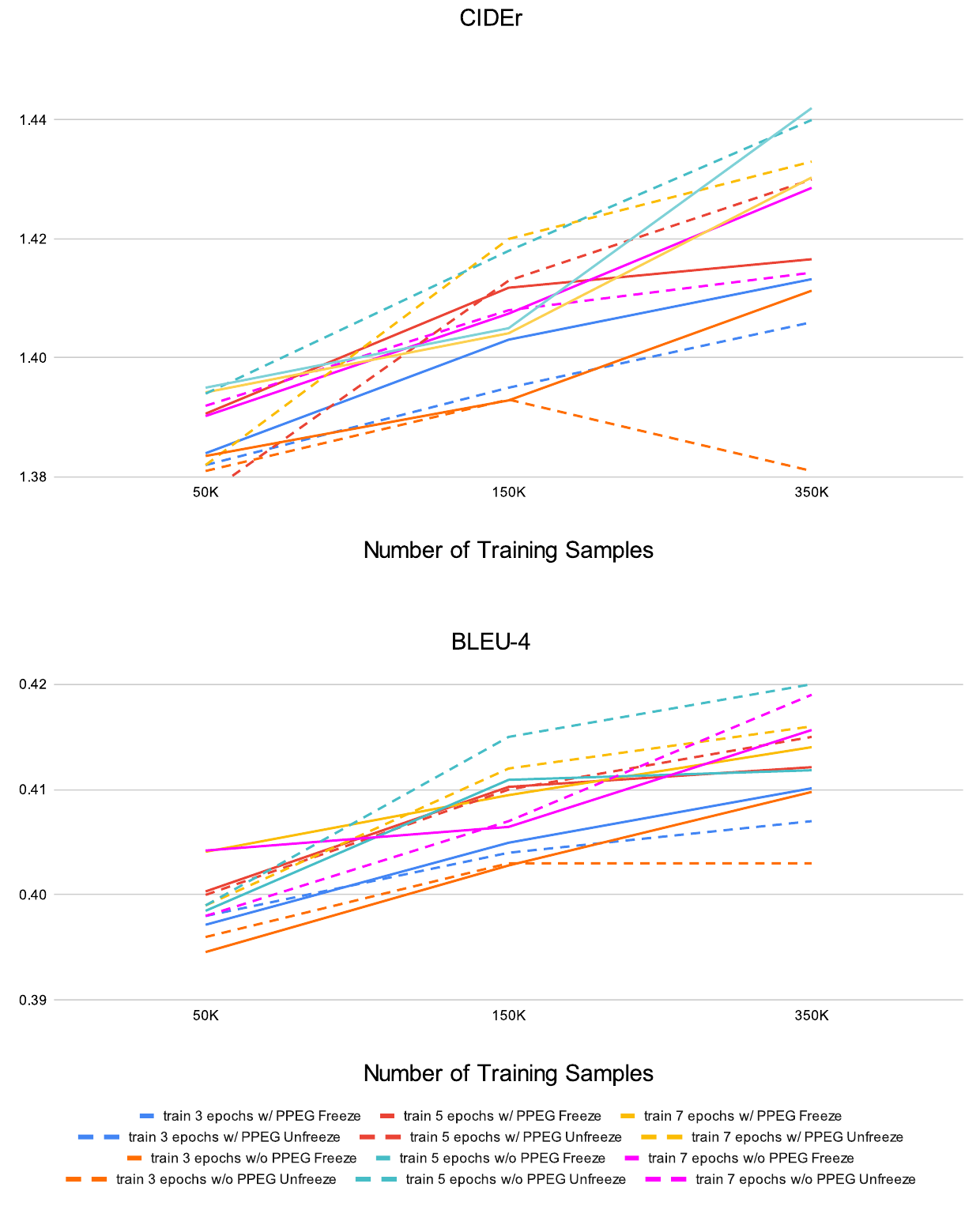}
    \caption{Experiment results on MSCOCO with or without freezing the visual encoder. We adopt the metrics used in \cite{li2023blip}.}
    \label{fig:unfreeze_ab}
\end{figure}

In this section, we validate the performance of fine-tuning while keeping the ViT frozen and image sizes unchanged. Experiment results can be seen as Figure \ref{fig:unfreeze_ab}. We observed that when working with limited data, such as 50K samples, models exhibit comparable performance whether or not the visual encoder (ViT) is frozen. However, as the number of training epochs increases, the performance gap varies. In some cases, unfreezing the ViT leads to improved performance, while in others, the opposite holds true. Considering that many real-world applications may not have access to massive training data, freezing the ViT can be a more efficient approach while still maintaining similar performance levels. 

\subsection{Case Study}
\label{sec:more_case_study}
In the main paper, we employ the FLAN-T5-XL as the language model. Existing large language models can be broadly categorized into two types: encoder-decoder based and decoder-only based models. The FLAN-T5-XL falls into the former category. The decoder-only based models are more computationally efficient and the encoder-decoder based models can handle more sophisticated tasks. In this section, we assess the performance of MIVPG on models from the decoder-only category. Specifically, we use the BLIP2\cite{li2023blip} with OPT-2.7b\cite{zhang2022opt} as the base LLM. We validate the performance on the PatchGastricADC22 dataset. In the experiments, we only replace the LLM while keeping other hyperparameters unchanged.

\begin{table}[ht]
    \centering
        \caption{Experiments on the PatchGastricADC22 dataset \cite{tsuneki2022inference} with OPT-2.7b as the language model}
        \label{tab:pga_results_opt}
        \resizebox{0.47\textwidth}{!}{
        \begin{tabular}{ccccc}
            \toprule
             & BLEU@4 & CIDEr & METEOR & ROUGE \\
             \midrule
             BLIP2-MIVPG w/o CSA & 0.427$\pm 0.012$ & 3.12$\pm 0.118$ & \textbf{0.349}$\pm 0.003$ & 0.494$\pm 0.123$\\
             BLIP2-MIVPG & \textbf{0.432}$\pm0.025$ & \textbf{3.21}$\pm 0.105$ & 0.347$\pm 0.016$ & \textbf{0.569} $\pm 0.019$\\
            \bottomrule
        \end{tabular}}
\end{table}

The experiment results on PatchGastricADC22 using OPT-2.7b as the language model are presented in Table \ref{tab:pga_results_opt}. Overall, the model continues to outperform the baselines shown in Table \ref{tab:pga_results}, emphasizing the advantages of integrating MLLMs into the WSI captioning task. Notably, the model with CSA performs better than the one without it, reaffirming the effectiveness of CSA. It's also worth noting that the performance of using OPT-2.7b is not superior to using Flan-T5-XL. This could be attributed, in part, to the insufficiency of training data. Since OPT-2.7b is relatively less sophisticated, more training data may be required to train a more powerful model.


\subsection{More Visualization}
This section provides additional visualization results on the ABO dataset, including both patch-level attention weights and image-level attention weights. In the patch-level attention weights, it is evident that the model excels in detecting the shapes of objects, as a significant portion of the patch-level weights is assigned to edges and contours. The image-level attention weights display maps for all twelve heads. Each row in a map represents a query, while each column represents an image. It's important to note that different heads and queries exhibit varying attention patterns towards the images, demonstrating the diversity in how the model processes and attends to the input images.

\label{sec:more_visualization}
\begin{figure*}[!h]
    \centering
    \includegraphics[width=1.\textwidth]{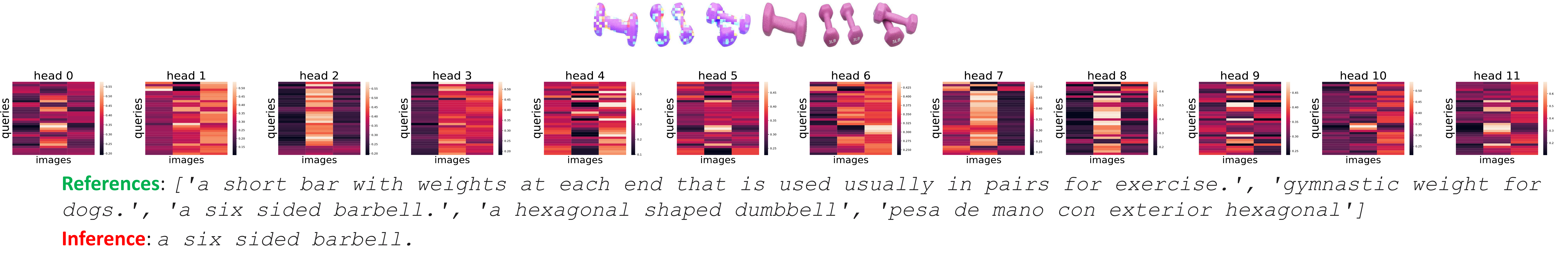}
\end{figure*}
\begin{figure*}[!h]
    \centering
    \includegraphics[width=1.\textwidth]{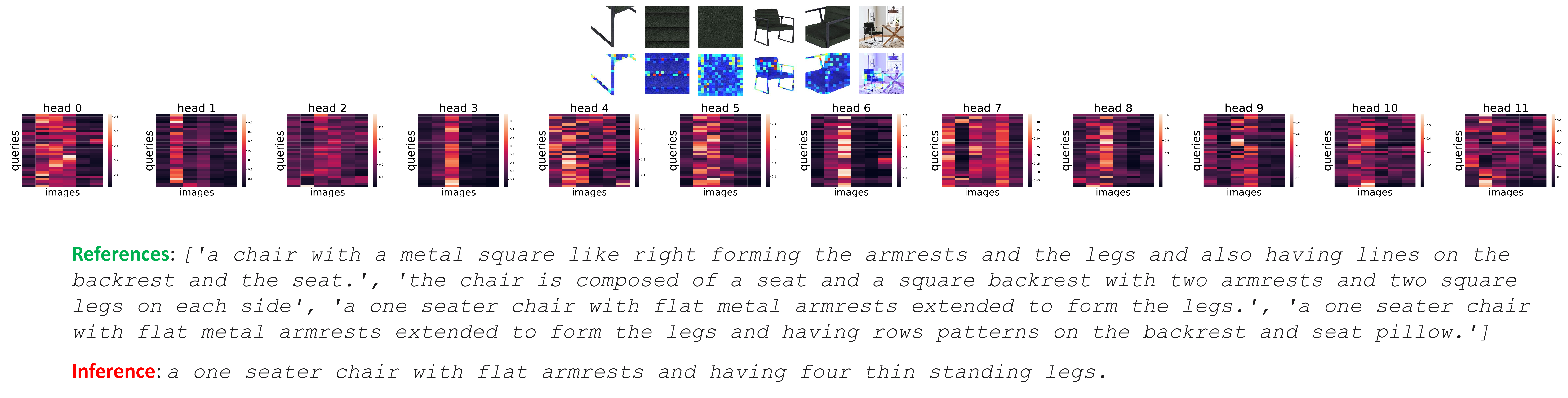}
\end{figure*}
\begin{figure*}[!h]
    \centering
    \includegraphics[width=1.\textwidth]{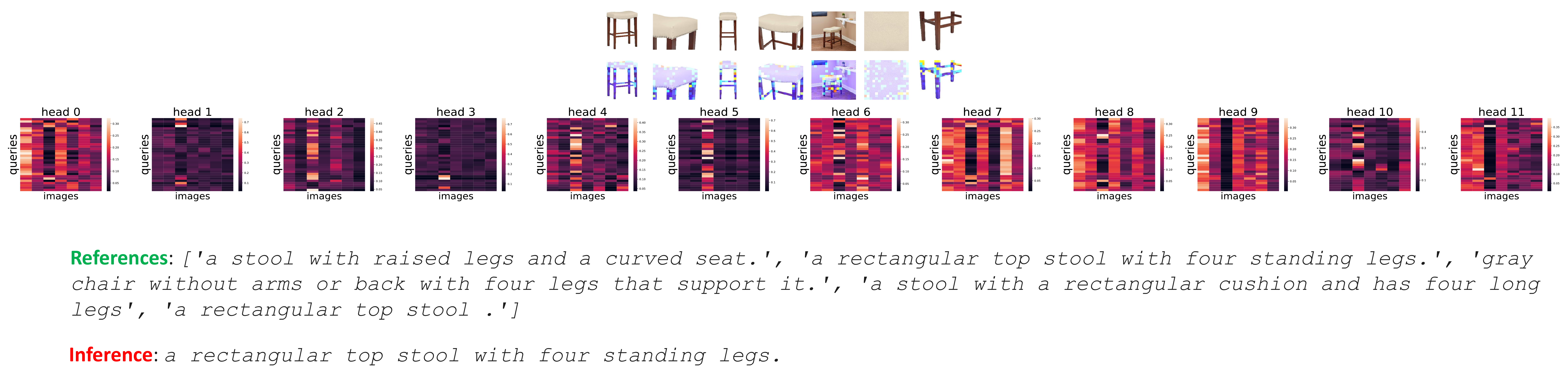}
\end{figure*}
\begin{figure*}[!h]
    \centering
    \includegraphics[width=1.\textwidth]{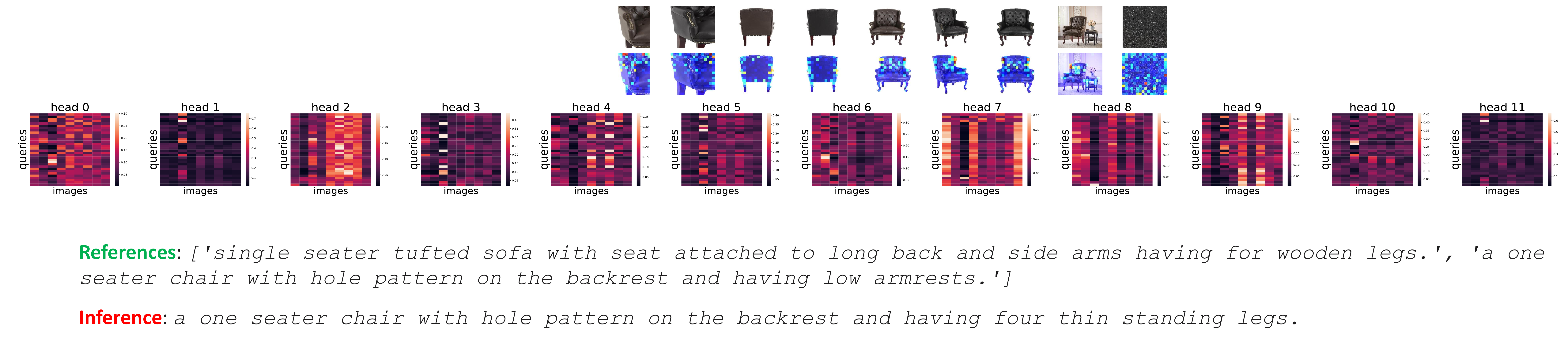}
\end{figure*}
\begin{figure*}[!h]
    \centering
    \includegraphics[width=1.\textwidth]{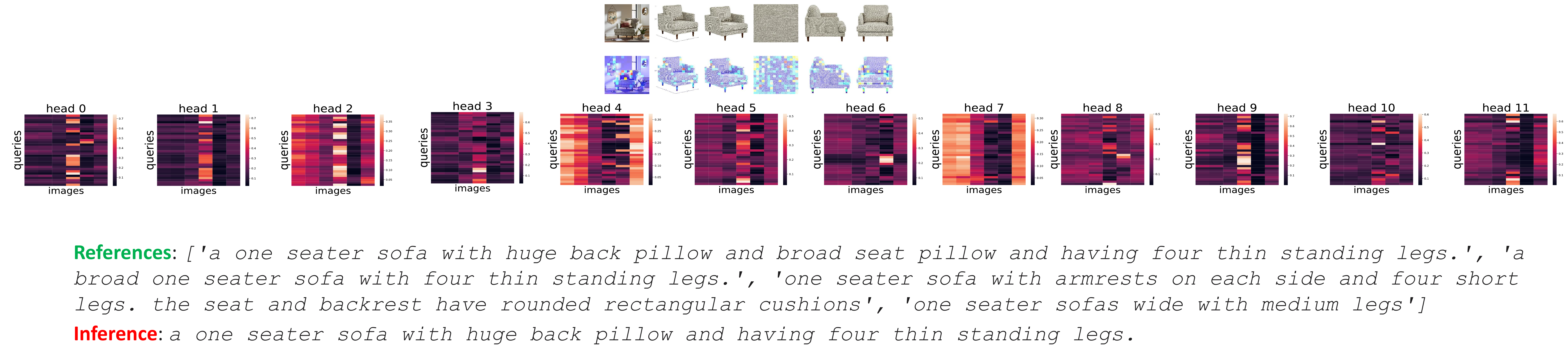}
\end{figure*}\begin{figure*}[!h]
    \centering
    \includegraphics[width=1.\textwidth]{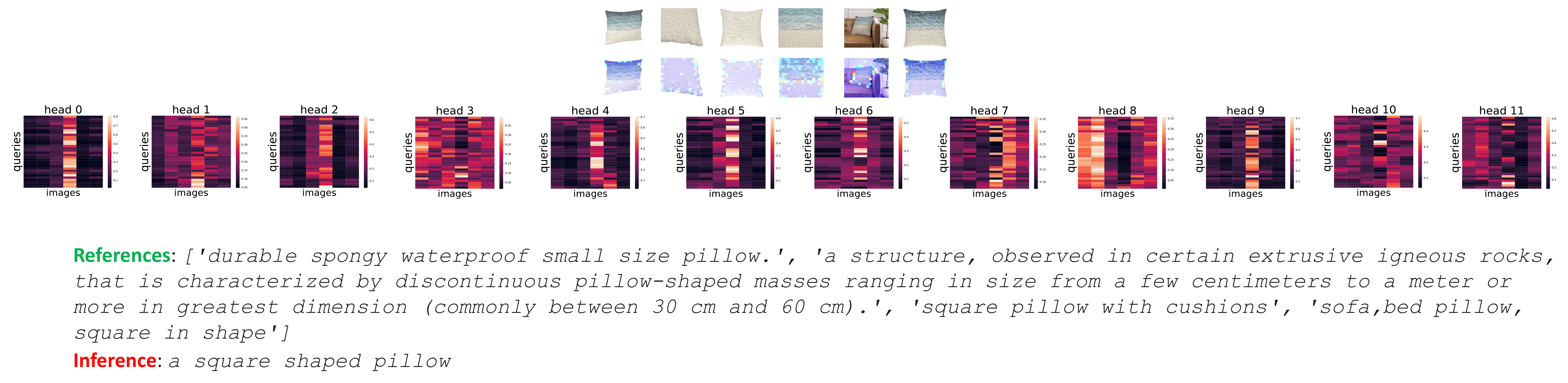}
\end{figure*}
\begin{figure*}[!h]
    \centering
    \includegraphics[width=1.\textwidth]{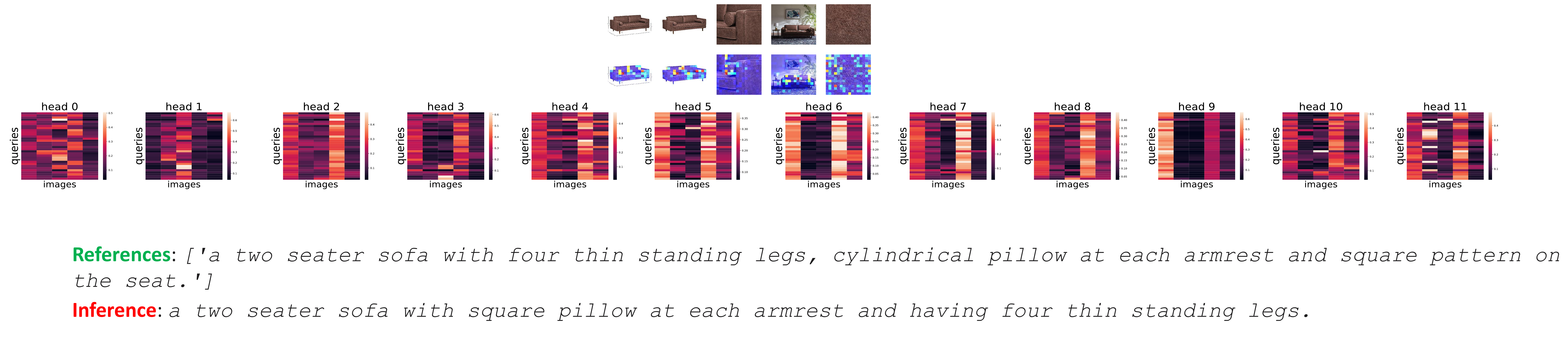}
\end{figure*}
\begin{figure*}[!h]
    \centering
    \includegraphics[width=1.\textwidth]{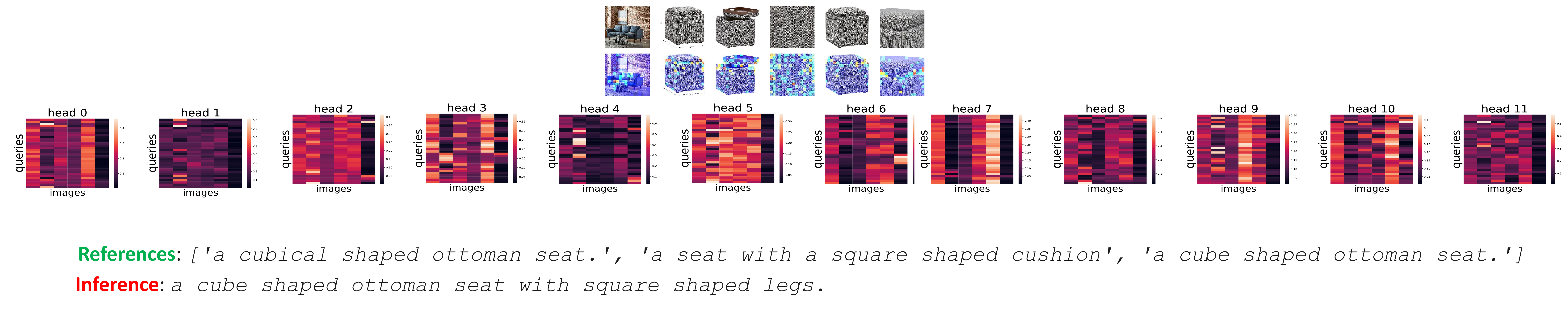}
    \label{fig:vis_8}
\end{figure*}
\begin{figure*}[!h]
    \centering
    \includegraphics[width=1.\textwidth]{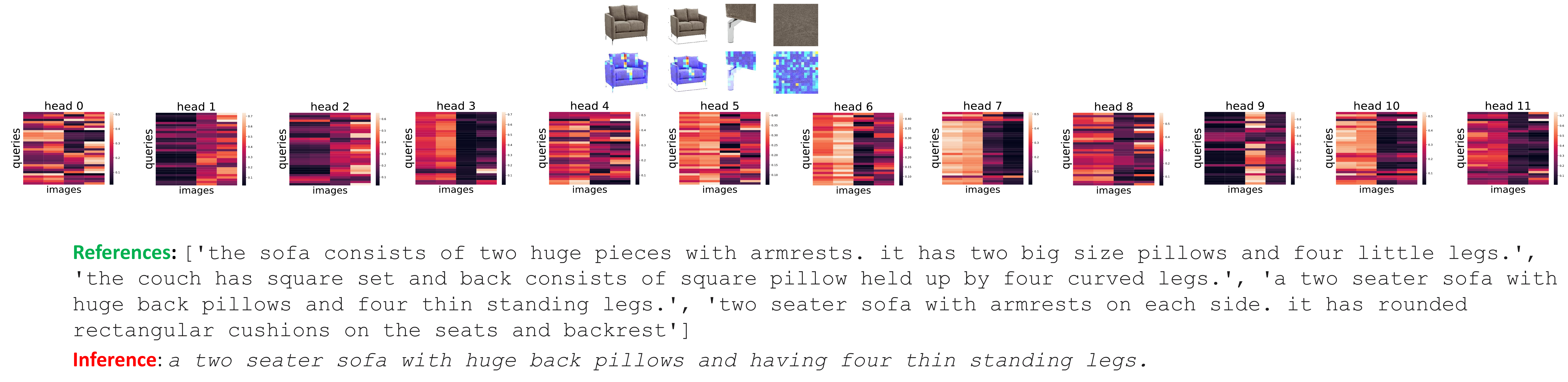}
\end{figure*}
\begin{figure*}[!h]
    \centering
    \includegraphics[width=1.\textwidth]{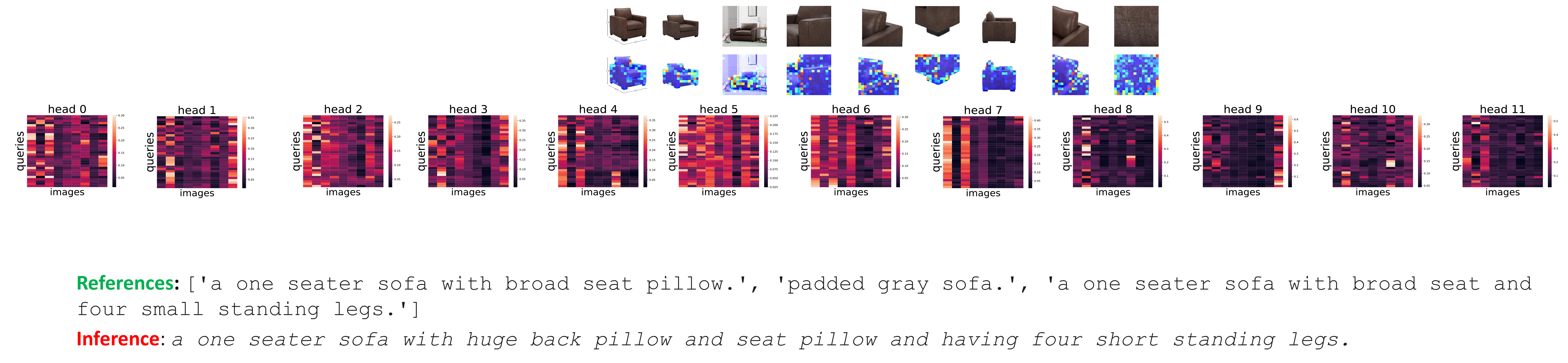}
\end{figure*}

\end{document}